\documentclass[12pt]{article}

\usepackage{color}
\usepackage[T1]{fontenc}    
\usepackage{url}            
\usepackage{booktabs}       
\usepackage{amsfonts}       
\usepackage{nicefrac}       
\usepackage{microtype}      
\usepackage{microtype}
\usepackage{graphicx}
\usepackage{subfigure}
\usepackage{booktabs} 
\usepackage{amsfonts}
\usepackage{bbm}
\usepackage{amssymb,amsmath,amsthm}
\usepackage{amsmath}
\usepackage{amsfonts}
\usepackage{mathrsfs}
\usepackage{amssymb}
\usepackage{amsthm}
\usepackage{wrapfig}
\newtheorem{definition}{Definition}

\usepackage{thmtools}
\usepackage{thm-restate}
\usepackage{hyperref}

\usepackage{cleveref}

\usepackage{graphicx,subfigure}
\usepackage{tikz}
\usepackage{comment}

\usepackage{epstopdf}

\usepackage{libertine}

\usepackage{algorithm}
\usepackage{algorithmic}

\usepackage[numbers]{natbib}
\newdimen\arrowsize
\pgfarrowsdeclare{arcsq}{arcsq}
{
  \arrowsize=0.2pt
  \advance\arrowsize by .5\pgflinewidth
  \pgfarrowsleftextend{-4\arrowsize-.5\pgflinewidth}
  \pgfarrowsrightextend{.5\pgflinewidth}
}
{
  \arrowsize=1.5pt
  \advance\arrowsize by .5\pgflinewidth
  \pgfsetdash{}{0pt} 
  \pgfsetroundjoin   
  \pgfsetroundcap    
  \pgfpathmoveto{\pgfpoint{0\arrowsize}{0\arrowsize}}
  \pgfpatharc{-90}{-140}{4\arrowsize}
  \pgfusepathqstroke
  \pgfpathmoveto{\pgfpointorigin}
  \pgfpatharc{90}{140}{4\arrowsize}
  \pgfusepathqstroke
}

\newtheorem{corollary}{Corollary}

\usepackage{chngcntr}
\usepackage{apptools}
\AtAppendix{\counterwithin{theorem}{section}}
\newtheorem{theorem}{Theorem}
\AtAppendix{\counterwithin{lemma}{section}}
\newtheorem{lemma}{Lemma}
\AtAppendix{\counterwithin{equation}{section}}

\title{An Information-Theoretic View for Deep Learning}

\font\myfont=cmr12 at 13pt

\author{\myfont Jingwei~Zhang\thanks{UBTECH Sydney AI Centre and the School of Information Technologies in the Faculty of Engineering and Information Technologies at The University of Sydney, NSW, 2006, Australia, zjin8228@uni.sydney.edu.au, tongliang.liu@sydney.edu.au, dacheng.tao@sydney.edu.au.} \ \ \   Tongliang~Liu\footnotemark[1] \ \ \    Dacheng~Tao\footnotemark[1]}

\date{}

\begin{document}

\newcommand*\sqcitep[1]{{\setcitestyle{square}$\!\!$\citep{#1}}}

\maketitle

\begin{abstract}
Deep learning has transformed computer vision, natural language processing, and speech recognition\cite{badrinarayanan2017segnet, dong2016image, ren2017faster, ji20133d}. However, two critical questions remain obscure: (1) why do deep neural networks generalize better than shallow networks; and (2) does it always hold that a deeper network leads to better performance? Specifically, letting $L$ be the number of convolutional and pooling layers in a deep neural network, and $n$ be the size of the training sample, we derive an upper bound on the expected generalization error for this network, i.e.,
 \begin{eqnarray*}
 \mathbb{E}[R(W)-R_S(W)]  \leq \exp{\left(-\frac{L}{2}\log{\frac{1}{\eta}}\right)}\sqrt{\frac{2\sigma^2}{n}I(S,W) }
 \end{eqnarray*} where $\sigma >0$ is a constant depending on the loss function, $0<\eta<1$ is a constant depending on the information loss for each convolutional or pooling layer, and $I(S, W)$ is the mutual information between the training sample $S$ and the output hypothesis $W$. This upper bound shows that as the number of convolutional and pooling layers $L$ increases in the network, the expected generalization error will decrease exponentially to zero. Layers with strict information loss, such as the convolutional layers, reduce the generalization error for the whole network; this answers the first question. However, algorithms with zero expected generalization error does not imply a small test error or $\mathbb{E}[R(W)]$. This is because $\mathbb{E}[R_S(W)]$ is large when the information for fitting the data is lost as the number of layers increases. This suggests that the claim ``the deeper the better'' is conditioned on a small training error or $\mathbb{E}[R_S(W)]$. Finally, we show that deep learning satisfies a weak notion of stability and the sample complexity of deep neural networks will decrease as $L$ increases.  
\end{abstract}
\newpage

\section{Introduction}

We study the standard statistical learning framework, where the instance space is denoted by $\mathcal{Z}$ and the hypothesis space is denoted by $\mathcal{W}$. The training sample is denoted by $S=\{Z_1,Z_2,...,Z_n\}$, where each element $Z_i$ is drawn i.i.d. from an unknown distribution $D$. A learning algorithm $\mathcal{A}: S \rightarrow \mathcal{W}$ can be regarded as a randomized mapping from the training sample space $\mathcal{Z}^n$ to the hypothesis space $\mathcal{W}$. The learning algorithm $\mathcal{A}$ is characterized by a Markov kernel $P_{W|S}$, meaning that, given training sample $S$, the algorithm picks a hypothesis in $\mathcal{W}$ according to the conditional distribution $P_{W|S}$.

We introduce a loss function $\ell: \mathcal{W}\times Z \to \mathbb{R}^{+}$ to measure the quality of a prediction w.r.t. a hypothesis.
For any learned hypothesis $W$ by $S$, we define the expected risk
 \begin{equation}
 R(W)=\mathbb{E}_{Z\sim D}[\ell(W, Z)]~,
 \end{equation}
 and the empirical risk
 \begin{equation}
 R_S(W)=\frac{1}{n}\sum_{i=1}^{n}\ell(W,Z_i)~.
 \end{equation}
For a learning algorithm $\mathcal{A}$, the generalization error is defined as
\begin{equation}
G_S(D, P_{W|S}) = R(W)-R_S(W)~.
\end{equation}
A small generalization error implies that the learned hypothesis will have similar performances on both the training and test datasets.
 
In this paper, we study the following expected generalization error for deep learning:
\begin{equation}\label{eq_gen}
G(D, P_{W|S}) = \mathbb{E}[R(W)-R_S(W)]~,
\end{equation}
where the expectation is over the joint distribution $P_{W,S}=D^n \times P_{W|S}$.

We have the following decomposition:
\begin{eqnarray}
\mathbb{E}[R(W)]=G(D, P_{W|S}) +\mathbb{E}[R_S(W)]~,
\end{eqnarray}
where the first term on the right-hand side is the expected generalization error, and the second term reflects how well the learned hypothesis fits the training data from an expectation view.

When designing a learning algorithm, we want the expectation of the expected risk, i.e., $\mathbb{E}[R(W)]$, to be as small as possible. However, obtaining small values for the expected generalization error $G(D, P_{W|S})$ and the expected empirical risk $\mathbb{E}[R_S(W)]$ at the same time is difficult. Usually, if a model fits the training data too well, it may generalize poorly on the test data; this is known as the bias-variance trade-off problem \cite{domingos2000unified}.
Surprisingly, deep learning has empirically shown their power for simultaneously minimizing $G(D, P_{W|S})$ and $\mathbb{E}[R_S(W)]$. They have small $\mathbb{E}[R_S(W)]$ because neural networks with deep architectures can efficiently compactly represent highly-varying functions \cite{sonoda2015neural}.
However, the theoretical justification for their small expected generalization errors $G(D, P_{W|S})$ remains elusive.

In this paper, we study the expected generalization error for deep learning from an information-theoretic point of view. We will show that, as the number of layers grows, the expected generalization error $G(D, P_{W|S})$ decreases exponentially to zero\footnote{
We have $I(S,W) \leq H(S)$, which is independent of $L$. Detailed discussions will be in Section \ref{section4} and Section \ref{learnability}~.}. Specifically, in Theorem \ref{main}, we prove that
 \begin{eqnarray*}
 &&G(D, P_{W|S})=\mathbb{E}[R(W)-R_S(W)] \\
 && \leq \exp{\left(-\frac{L}{2}\log{\frac{1}{\eta}}\right)}\sqrt{\frac{2\sigma^2}{n}I(S,W) }~,
 \end{eqnarray*}
where $L$ is the number of information loss layers in deep neural networks (DNNs), $0<\eta<1$ is a constant depending on the average information loss of each layer, $\sigma >0$ is a constant depending on the loss function, $n$ is the size of the training sample $S$, and $I(S, W)$ is the mutual information between the input training sample $S$ and the output hypothesis $W$. The advantage of using the mutual information between the input and output to bound the expected generalization error \cite{2015arXiv151105219R,NIPS2017_6846} is that it depends on almost every aspects of the learning algorithm, including the data distribution, the complexity of the hypothesis class, and the property of the learning algorithm itself.

Our result is consistent with the bias-variance trade-off. Although the expected generalization error decreases exponentially to zero as the number of information loss layers increases, the expected empirical risk $\mathbb{E}[R_S(W)]$ increases since the information loss is harmful to data fitting. This implies that, when designing deep neural networks, greater efforts should be made to balance the information loss and expected training error.

We also provide stability and risk bound analyses for deep learning. We prove that deep learning satisfies a weak notion of stability, which we term average replace-one hypothesis stability, implying that the output hypothesis will not change too much by expectation when one point in the training sample is replaced. Under the assumption that the algorithm mapping is deterministic, the notion of average replace-one hypothesis stability will degenerate to the case of average replace-one stability, as proposed by \cite{shalev2010learnability}, which has been identified as a necessary condition for learnability in the general learning setting introduced by Vapnik. 

We further provide an expected excess risk bound for deep learning and show that the sample complexity of deep learning will decrease as $L$ increases, which surprisingly indicates that by increasing $L$, we need a smaller sample complexity for training. However, this does not imply that increasing the number of layers will always help. An extreme case is that, as $L$ goes to infinity, the output feature will lose all predictive information and no training sample is needed because random-guessing is optimal. We also derive upper bounds of the expected generalization error for some specific deep learning algorithms, such as noisy stochastic gradient decent (SGD) and binary classification for deep learning. We further show that these two algorithms are PAC-learnable with sample complexities of $\widetilde{\mathcal{O}}(\frac{1}{\sqrt{n}})$.

The remainder of this paper is organized as follows. In Section \ref{section2}, we relate DNNs to Markov chains. Section \ref{section3} exploits the strong data processing inequality to derive how the mutual information, between intermediate features representations and the output, varies in DNNs. Our main results are given in Section \ref{section4}, which gives an exponential generalization error bound for DNNs in terms of the depth $L$; we then analyze the stability of deep learning in Section \ref{stability} and the learnability for deep learning with noisy SGD and binary classification in Section \ref{learnability}; Section \ref{discuss} makes some discussions; all the proofs are provided in Section \ref{proofs}; finally, we conclude our paper and highlight some important implications in Section \ref{section5}~.

\section{The Hierarchical Feature Mapping of DNNs and Its Relationship to Markov Chains } \label{section2}
\begin{figure} 
  \centering
    \includegraphics[width=1\textwidth]{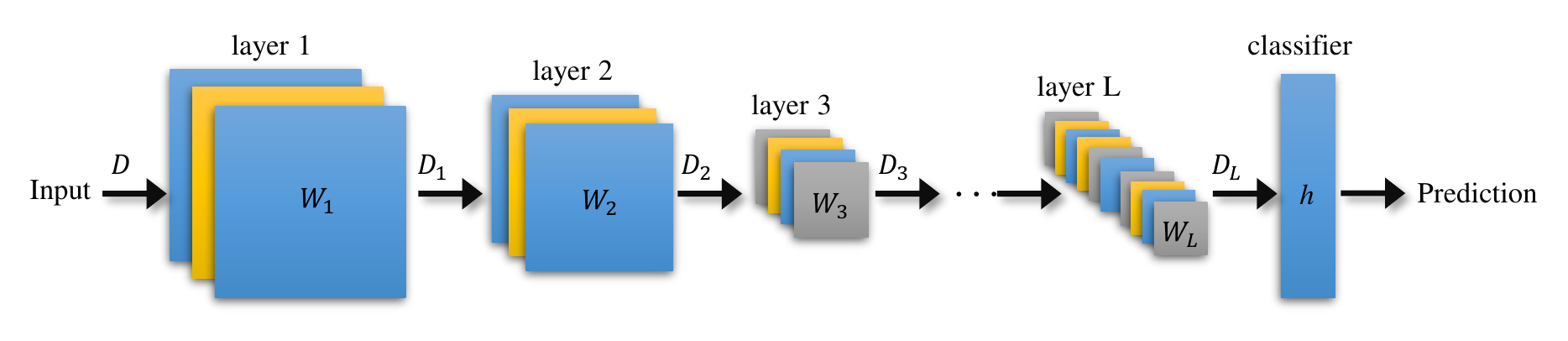}
    \caption{Hierarchical Feature Mapping of Deep Neural Networks with L Hidden Layers}
      \label{fig1}
\end{figure}

\begin{figure} 
  \centering
    \includegraphics[width=0.7\textwidth]{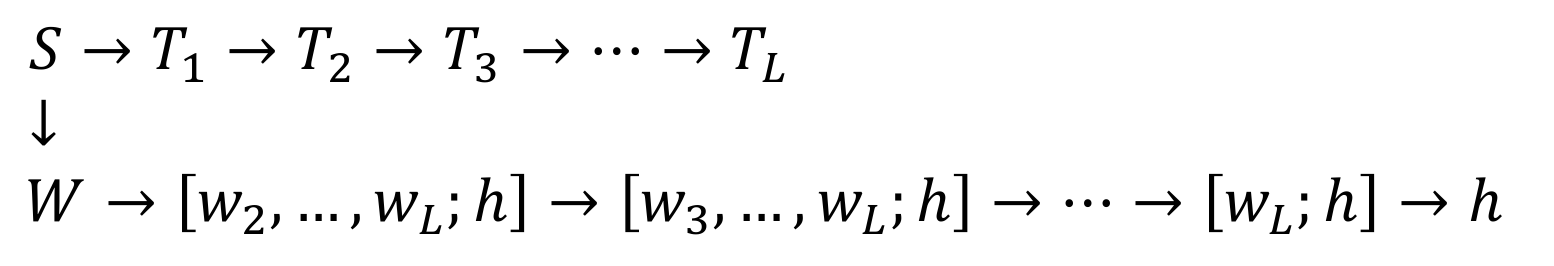}
      \caption{The Feature Mapping of Deep Neural Networks Forms a Markov Chain, when given $w_1,\ldots,w_L$.}
       \label{fig2}
\end{figure}
We first introduce some notations for deep neural networks (DNNs).
As shown in Figure \ref{fig1}, a DNN with $L$ hidden layers can be seen as $L$ feature maps that sequentially conduct feature transformations $L$ times on the input $Z$. After $L $ feature transformations, the learned feature will be the input of a classifier (or regressor) at the output layer. If the distribution on a single input is $D$, then we denote the distribution after going through the $k$-th hidden layer as $D_k$ and the corresponding variable as $\widetilde{Z}_k$ where $k=1,\ldots,L$. The weight of the whole network is denoted by $W=[w_1,\ldots,w_L;h]\in\mathcal{W}$, where $\mathcal{W}$ is the space of all possible weights. As shown in Figure \ref{fig2}, the input $S$ is transformed layer by layer and the output of the $k$-th hidden layer is $T_k$, where $k=1,\ldots,L$. We also denote the $j$-th sample after going through the $k$-th hidden layer by $Z_{k_j}$. In other words, we have the following relationships:
\begin{eqnarray}
&& Z \sim D, \\
&& \widetilde{Z}_k \sim D_k,\quad for \quad k =1,\ldots,L, \\
&& S=\{Z_1, \ldots,Z_n\}\sim D^n,\\
&& T_k = \{Z_{k_1},\ldots,Z_{k_n}\}\sim D_k^n, \nonumber \\
&& when ~~given~~ w_1,\ldots,w_{k}, \quad for\quad k=1, \ldots, L.
\end{eqnarray}

We now have a Markov model for DNNs, as shown in Figure \ref{fig2}. From the Markov property, we know that if $U\rightarrow V \rightarrow W$ forms a Markov chain, then $W$ is conditionally independent of $U$ given $V$. Furthermore, from the data processing inequality \cite{cover2012elements}, we have $I(U, W)\leq I(U,V)$, and the equality holds if and only if $U\rightarrow W \rightarrow V$ also forms a Markov chain. Applying the data processing inequality to the Markov chain in Figure \ref{fig2}, we have,
\begin{eqnarray} \label{dpi}
&&I(T_L, h|w_1,\ldots,w_L) \leq I(T_{L-1}, h|w_1,\ldots,w_L) \nonumber\\
&&\leq I(T_{L-2}, h|w_1,\ldots,w_L)\leq \ldots \leq I(S, h|w_1,\ldots,w_L)\nonumber\\
&& =I(S, W|w_1,\ldots,w_L)~.
\end{eqnarray}
This means that the mutual information between input and output is non-increasing as it goes through the network layer by layer. As the feature map in each layer is likely to be non-invertible, the mutual information between the input and output is likely to strictly decrease as it goes through each layer. This encourages the study of the strong data processing inequality \cite{2015arXiv150806025P, ahlswede1976spreading}. In the next section, we prove that the strong data processing inequality holds for DNNs in general.

\section{Information Loss in DNNs}\label{section3}
In the previous section, we model a DNN as a Markov chain and conclude that the mutual information between input and output in DNNs is non-increasing by using the data processing inequality. The equalities in equation (\ref{dpi}) will not hold for most cases because the feature mapping is likely to be non-invertible, and therefore we can apply the strong data processing inequality to achieve tighter inequalities.

For a Markov chain $U\rightarrow V\rightarrow W$, the random transformation $P_{W|V}$ can be seen as a channel from an information-theoretic point of view.
Strong data processing inequalities (SDPIs) quantify an intuitive observation that the noise inside channel $P_{W|V}$ will reduce the mutual information between $U$ and $W$. That is, there exists $0\leq\eta< 1$, such that
\begin{eqnarray}
I(U,W)\leq \eta I(U,V)~.
\end{eqnarray}
Formally,
\begin{theorem}\cite{ahlswede1976spreading} \label{lemma1}
Consider a Markov chain $W\rightarrow X\rightarrow Y$ and the corresponding random mapping $P_{Y|X}$. If the mapping $P_{Y|X}$ is noisy (that is, we cannot recover $X$ perfectly from the observed random variable $Y$), then there exists $0\leq\eta<1$, such that
\begin{eqnarray}
I(W,Y)\leq \eta I(W,X)
\end{eqnarray}
\end{theorem}

More details can be found in a comprehensive survey on SDPIs \cite{2015arXiv150806025P}.

Let us consider the $k$-th hidden layer ($1\leq k\leq L$) in Figure \ref{fig1}. This can be seen as a randomized transformation $P_{\widetilde{Z}_{k}|\widetilde{Z}_{k-1}}$ mapping from one distribution $D_{k-1}$ to another distribution $ D_k$ (when $k=1$, we denote $D=D_0$ ). We then denote the parameters of the $k$-th hidden layer by $w_k$\footnote{The bias for each layer can be included in $w_k$ via homogeneous coordinates.}. Without loss of generality, let $w_k$ be a matrix in $  \mathbb{R}^{d_{k}\times d_{k-1}}$. Also, we denote the activation function in this layer by $\sigma_k(\cdot)$.
\begin{definition}[Contraction Layer]
A layer in a deep neural network is called a contraction layer if it causes information loss.
\end{definition}
We now give the first result, which quantifies the information loss in DNNs.
\begin{corollary}[Information Loss in DNNs] \label{thm1}
Consider a DNN as shown in Figure \ref{fig1} and its corresponding Markov model in Figure \ref{fig2}. If its $k$-th ($1\leq k\leq L$ ) hidden layer  is a contraction layer, then there exists $0\leq \eta_k <1$, such that
\begin{eqnarray}
I(T_k, h|w_1,\ldots,w_L)\leq \eta_k I(T_{k-1}, h|w_1,\ldots,w_L)~.
\end{eqnarray}
\end{corollary}
We show that the most used convolutional and pooling layers are contraction layers.
\begin{lemma}[proved in 8.1]\label{lemma2}
For any layer in a DNN, with parameters $w_k\in  \mathbb{R}^{d_{k}\times d_{k-1}}$, if $rank(w_k)< d_{k-1}$, it is a contraction layer.
\end{lemma}
Corollary \ref{thm1} shows that the mutual information $I(T_{k-1}, h|w_1,\ldots,w_L)$ decreases after it goes through a contraction layer. From Lemma \ref{lemma2}, we know that the convolutional and pooling layers are guaranteed to be contraction layers. Besides, when the shape of the weight $w_k$ satisfies $d_{k}<d_{k-1}$, it also leads to a contraction layer.  For a fully connected layer with shape $d_{k}\geq d_{k-1}$, the contraction property sometimes may not hold because when the weight matrix is of full column rank with probability $1$, it will leads to a noiseless and invertible mapping. However, the non-invertible activation function  (e.g. ReLU activation) employed sub-sequentially can contribute to forming a contraction layer.
Without loss of generality, in this paper, we let all $L$ hidden layers be contraction layers, e.g., convolutional or pooling layers.

\section{Exponential Bound on the Generalization Error of DNNs}\label{section4}
Before we introduce our main theorem, we need to restrict the loss function $\ell(W,Z)$ to be $\sigma$-sub-Gaussian with respect to $(W,Z)$ given any $w_1,\ldots,w_L$.
\begin{definition}[$\sigma$-sub-Gaussian]
A random variable X is said to be $\sigma$-sub-Gaussian if the following inequality holds for any $\lambda\in \mathbb{R}$,
\begin{eqnarray}
\mathbb{E}[\exp\left(\lambda(X-\mathbb{E}[X])\right)] \leq \exp\left(\frac{\sigma^2\lambda^2}{2}\right)~.
\end{eqnarray}
\end{definition}

We now present our main theorem, which gives an exponential bound for the expected generalization error of deep learning.
\begin{theorem}[proved in 8.2] \label{main}
For a DNN with $L$ hidden layers, input $S$, and parameters $W$, assume that the loss function $\ell(W,Z)$ is $\sigma$-sub-Gaussian with respect to $(W, Z)$ given any $w_1,\ldots,w_L$. Without loss of generality, let all $L$ hidden layers be contraction layers. Then, the expected generalization error can be upper bounded as follows,
 \begin{eqnarray} \label{eqa}
 \mathbb{E}[R(W)-R_S(W)]  \leq \exp{\left(-\frac{L}{2}\log{\frac{1}{\eta}}\right)}\sqrt{\frac{2\sigma^2}{n}I(S,W) }
 \end{eqnarray}
 where $\eta<1$ is the geometric mean of information loss factors of all L contraction layers, that is
  \begin{eqnarray}
 \eta=\left(\mathbb{E}_{w_1,\ldots,w_L}\left(\prod_{k=1}^{L}\eta_k\right)\right)^{\frac{1}{L}} ~.
 \end{eqnarray}
\end{theorem}

The upper bound in Theorem \ref{main} may be loose w.r.t. the mutual information $I(S,W)$ since we used the inequality $I_{cond}(S,W|w_1,\ldots,w_L)\leq I(S,W)$ in the proof. We also have that
\begin{eqnarray}
I(S,W) \leq H(S)~.
\end{eqnarray}
By definition, $\eta<1$ holds uniformly for any given $L$ and $\eta^L$ is a strictly decreasing function of $L$. These imply that as the number of contraction layers $L$ increases, the expected generalization error will decrease exponentially to zero.

Theorem \ref{main} implies that deeper neural networks will improve the generalization error. However, this does not mean that the deeper the better. Recall that $\mathbb{E}[R(W)]=G(D, P_{W|S}) +\mathbb{E}[R_S(W)]$;  a small $G(D, P_{W|S})$ does not imply a small $\mathbb{E}[R(W)]$,
since the expected training error $\mathbb{E}[R_S(W)]$ increases due to information loss. Specifically, if the information about the relationship between the observation $X$ and the target $Y$ is lost, fitting the training data will become difficult and the expected training error will increase. Our results highlight a new research direction for designing deep neural networks, namely that we should increase the number of contraction layers while keeping the expected training error small.

Information loss factor $\eta$ plays an essential role in the generalization of deep learning. A successful deep learning model should filter out redundant information as much as possible while retaining sufficient information to fit the training data. The functions of some deep learning tricks, such as convolution, pooling, and activation, are very good at filtering out some redundant information. The implication behind our theorem somewhat coincides the information-bottleneck theory \cite{2017arXiv170300810S}, namely that with more contraction layers, more redundant information will be removed while predictive information is preserved.

\section{Stability and Risk Bound of Deep Learning} \label{stability}
It is known that the expected generalization error is equivalent to the notion of stability of the learning algorithm \cite{shalev2010learnability}. In this section, we show that deep learning satisfies a weak notion of stability and, further, show that it is a necessary condition for the learnability of deep learning. We first present a definition of stability, as proposed by \cite{shalev2010learnability}. 
\begin{definition}\cite{shalev2010learnability}
A learning algorithm $\mathcal{A}: S\to \mathcal{W}$ is \emph{average replace-one stable} with rate $\alpha(n)$ under distribution $D$ if
\begin{eqnarray}
\left | \frac{1}{n}\sum_{i=1}^{n}\mathbb{E}_{S\sim D^n,Z_i^{\prime}\sim D}\left[\ell(W,Z_i^{\prime})-\ell(W^i,Z_i^{\prime})\right] \right| \leq \alpha(n)~.
\end{eqnarray}
\end{definition}
For deep learning, we define another notion of stability, that we term \emph{average replace-one hypothesis stability}. 
\begin{definition}[average replace-one hypothesis stability]
A learning algorithm $\mathcal{A}: S\to \mathcal{W}$ is \emph{average replace-one hypothesis stable} with rate $\beta(n)$ under distribution $D$ if
\begin{eqnarray}
&&\left | \frac{1}{n}\sum_{i=1}^{n}\mathbb{E}_{S\sim D^n,Z_i^{\prime}\sim D, W\sim P_{W|S}, W^i\sim P_{W^i|S^i}}\left[\ell(W,Z_i^{\prime})-\ell(W^i,Z_i^{\prime})\right]\right|\leq \beta(n)~.
\end{eqnarray}
\end{definition}
The difference between average replace-one hypothesis stability and average replace-one stability is that the former one also takes an expectation over $W\sim P_{W|S}$ and $W^i\sim P_{W^i|S^i}$, which is weaker than average replace-one stability.
It can clearly be seen that average replace-one stability with rate $\alpha(n)$ implies average replace-one hypothesis stability with rate $\alpha(n)$. We now prove that deep learning is  average replace-one hypothesis stable.
\begin{theorem}[proved in 8.3] \label{thm3}
Deep learning is average replace-one hypothesis stable with rate 
\begin{eqnarray}
\beta(n)=\exp{\left(-\frac{L}{2}\log{\frac{1}{\eta}}\right)}\sqrt{\frac{2\sigma^2}{n}I(S,W) }~.
\end{eqnarray}
\end{theorem}

Deep learning algorithms are average replace-one hypothesis stable, which means that replacing one training example does not alter the output too much as shown in Theorem \ref{thm3}.

 As concluded by \cite{shalev2010learnability}, the property of average replace-one stability is a necessary condition for characterizing learnability. We have also shown that average replace-one stability implies average replace-one hypothesis stability. Therefore, the property of average replace-one hypothesis stability is a necessary condition for the learnability of deep learning. However, it is not a sufficient condition. Finding a necessary and sufficient condition for characterizing learnability for deep learning remains unsolved.
 
\section{Learnability, Sample Complexity, and Risk Bound for Deep Learning}  \label{learnability}
We have derived an exponential upper bound of the expected generalization error for deep learning. In this section, we further derive the excess risk bound and analyze the sample complexity and learnability for deep learning in a general setting. We can roughly bound $I(S,W)$ by $H(S)$, which will be large when the input tends to be uniformly distributed.  Nevertheless, for some specific deep learning algorithms, a much tighter upper bound of the mutual information can be obtained. Here, we consider two cases where a tighter bound can be achieved. That is noisy SGD and binary classification in deep learning. We also derive the sample complexity for these two algorithms.

\subsection{Learnability and Risk Bound for Deep Learning} 
This subsection provides a qualitative analysis on the expected risk bound of deep learning. 
By picking any global expected risk minimizer,
\begin{eqnarray}
W^*=\arg\min_{W\in\mathcal{W}}R(W)
\end{eqnarray}
and picking any empirical risk minimizer
\begin{eqnarray} 
W =\arg\min_{W\in\mathcal{W}}R_S(W)~,
\end{eqnarray}
we have
\begin{eqnarray} \label{eqb}
&&\mathbb{E}_{W,S}[R_S(W)] \leq \mathbb{E}_{W, S}[R_S(W^*)]\nonumber\\
&&=\mathbb{E}_{S}[R_S(W^*)]=R(W^*)~.
\end{eqnarray}
Note that a global expected risk minimizer $W^*$ is neither dependent on $S$ nor a random variable, while $W$ is dependent on $S$. As mentioned before, we consider the case when $W$ is a random variable drawn according to the distribution $P_{W|S}$.

Therefore, by combining (\ref{eqa}) and (\ref{eqb}), we obtain an expected excess risk bound as follows,
\begin{eqnarray}  \label{eqc}
\mathbb{E}_{W, S}[R(W)] - R^* \leq \exp{\left(-\frac{L}{2}\log{\frac{1}{\eta}}\right)}\sqrt{\frac{2\sigma^2}{n}I(S,W) }
\end{eqnarray}
where $R^*=R(W^*)$.

It is worth noticing that $R^*$ is a non-decreasing function of $L$, because the rule constructed over the space $\widetilde{Z}_L$ cannot be better than the best possible rule in $\widetilde{Z}_{L-1}$, since all information in $\widetilde{Z}_L$ originates from space $\widetilde{Z}_{L-1}$. 
We now reach two conclusions:
\begin{itemize}
\item As the number of contraction layers $L$ goes to infinity, then both the excess risk and generalization error will decrease to zero. By strong data processing inequalities, $I(T_L, h|w_1,\ldots,w_L)$ will also decrease to zero\footnote{See 8.2 for more details.}, which means that the output feature $T_L$ will lose all predictive information. Therefore, no samples are needed for training, as any learned predictor over the transformed feature $T_L$ will perform no better than random guessing. In this case, although the sample complexity is zero, the optimal risk $R^*$ reaches its worst case.
\item As we increase the number of contraction layers $L$, the sample complexity will decrease. The result is surprising when $R^*$ is not increasing. This finding implies that if we could efficiently find a global empirical risk minimizer, we need smaller sample complexities when increasing the number of contraction layers. Besides, when these added contraction layers only filter out redundant information, $R^*$ will be not increasing. However, it is not easy to find the global empirical risk minimizer and control all contraction layers such that they only filter out redundant information. A promising new research direction is to increase the number of contraction layers while keeping a small $R^*$ or $\mathbb{E}[R_{S}(W)]$ or $R_{S}(W)$. 
\end{itemize}

We now discuss whether the deep learning is learnable in general. 
From equation (\ref{eqc}) and using Markov inequality, we have that with probability at least $1-\delta$,
\begin{equation} 
R(W) - R^* \leq \frac{1}{\delta}\exp{\left(-\frac{L}{2}\log{\frac{1}{\eta}}\right)}\sqrt{\frac{2\sigma^2}{n}I(S,W) } ~.
\end{equation}
We know that the notion of PAC-learnability in traditional learning theory must hold for any distribution $D$ over the instance space $\mathcal{Z}$. However, for the general case as presented in our main result, with different distribution $D$, an upper bound of the term $I(S,W)$ can vary and sometimes may be quite large even of the order $\mathcal{O}(n)$ (e.g. $I(S,W)\leq H(S)\leq n\log|\mathcal{X}||\mathcal{Y}|$). In this case, a sample complexity is $\mathcal{O}(\frac{1}{n^0})$, which is trivial and cannot guarantee the learnability as $n$ increases. In the next two subsections, we will show that for some specific deep learning algorithms, a tighter excess risk bound can be achieved and the sample complexity will be the order of $\widetilde{\mathcal{O}}(\frac{1}{\sqrt{n}})$.

\subsection{Generalization Error Bound With Noisy SGD in Deep Learning}
Consider  the problem of empirical risk minimization (ERM) via noisy mini-batch SGD in deep learning, where the weight $W$ is updated successively based on samples drawn from the training set $S$ and with a noisy perturbation.  The motivations of adding noise in SGD are mainly to prevent the learning algorithm from overfitting the training data and to avoid an exponential time to escape from saddle points \cite{NIPS2017_6707}. 

Denote the weight of a DNN at the time step $t$ by $W_t = [w_{1_t}, \ldots, w_{L_t}; h_t]$ and $\bold{Z}_t=\{Z_{t_1},\ldots,Z_{t_m}\}\subset S$ is the mini-batch with batch size $m$ at the $t$-th iteration\footnote{With some abuse of notations, $Z_{k_j}$ also denotes the $j$-th sample after going through the $k$-th hidden layer, where $k=1,\ldots,L$~, but it is not hard to distinguish them from the context.}. Then we have the updating rules 
$h_{t} = h_{t-1} -~ \alpha_t\left[\frac{1}{m}\sum_{i=1}^{m}\nabla_{h}\ell(W_{t-1}, Z_{t_i})\right]+n_t $ and 
$w_{k_{t}} = w_{k_{t-1}}- \beta_{k_t}\left[\frac{1}{m}\sum_{i=1}^{m}\nabla_{w_k}\ell(W_{t-1}, Z_{t_i})\right]+n_{k_t}$
where $k=1,\ldots, L$; $\alpha_t$ and $\beta_{k_t}$ denote the learning rates at the time step $t$ for each layer; $n_t \sim \mathcal{N}(0, \sigma_t^2 \bold{I}_{d})$ and $n_{k_t}\sim\mathcal{N}(0, \sigma_{k_t}^2 \bold{I}_{d_k}) $ are noisy terms that add a white Gaussian noise to each element of the update independently. Here, we assume that the updates of $h$ have bounded second moment. That is, there exists $0<M<\infty$, such that $\mathbb{E}\left[ \left|\left|\frac{1}{m}\sum_{i=1}^{m}\nabla_{h}\ell(W_{t-1}, Z_{t_i}) \right|\right|^2 \right] \leq M$ for all $t>0$. We have the following generalization error bound.
\begin{theorem} [proved in 8.4]\label{SGD}
For noisy SGD with bounded second moment in updates and $T$ iterations, the expected generalization error of deep learning can be upper bounded by
\begin{eqnarray}
&&\left |\mathbb{E}[R(W)-R_S(W)]\right|  \nonumber\\
&&\leq \exp{\left(-\frac{L}{2}\log{\frac{1}{\eta}}\right)}\sqrt{\frac{\sigma^2}{n}\sum_{i=1}^{T}\frac{M^2\alpha_i^2}{\sigma_i^2} } ~.
\end{eqnarray}
\end{theorem}
With the theorem above, we further prove the learnability and sample complexity of the noisy SGD in deep learning.
\begin{theorem} [proved and further discussed in 8.6]
The noisy SGD with bounded second moment in updates for deep learning is learnable, with the sample complexity of $\mathcal{O}\left(\frac{1}{\sqrt{n}}\right)$~.
\end{theorem}

\subsection{Generalization Error Bound for Binary Classification in Deep Learning}
This subsection gives an upper bound of the expected generalization error for deep learning in the case of binary classification.
For binary classification, we denote the function space of the classifier $h$ of the output layer by $\mathcal{H}$ and its VC-dimension by $\hat{d}$. The training set is $S=\{Z_1,\ldots,Z_n\}=\{(x_1, y_1),\ldots,(x_n,y_n)\}\in \mathcal{X}^n\times\mathcal{Y}^n$. When given $w_1,\ldots,w_L$, we have the transformed training set after $L$ feature mappings $T_L=\{(x_{L{_1}}, y_1),\ldots,(x_{L_{n}},y_n)\}\in\mathcal{X}_L^n\times\mathcal{Y}^n$ and $\mathcal{H}$ is a class of functions from $\mathcal{X}_L$ to $\{0,1\}$. For any integer $m\geq0$, we present the definition of the growth function of $\mathcal{H}$ as in \cite{mohri2012foundations}.
\begin{definition}[Growth Function]
The growth function of a function class $\mathcal{H}$ is defined as

\begin{equation}
\Pi_{\mathcal{H}}(m)=\max_{x_1,\ldots,x_m\in\mathcal{X}} \left| \{   (h(x_1),\ldots,h(x_m)):h\in\mathcal{H}   \} \right|~.
\end{equation}
\end{definition}

Now, we give a generalization error bound and a sample complexity for binary classification in deep learning in the following two theorems.

\begin{theorem}[proved in 8.5] \label{BINARY}
For binary classification in deep learning, the upper bound of the expected generalization error is given by 
\begin{eqnarray}
&& \left |\mathbb{E}[R(W)-R_S(W)]\right| \leq \exp{\left(-\frac{L}{2}\log{\frac{1}{\eta}}\right)}\sqrt{\frac{2\sigma^2\hat{d}}{n}}~~for~~n\leq \hat{d}
\end{eqnarray}
and

\begin{eqnarray}
&&\left |\mathbb{E}[R(W)-R_S(W)]\right|\leq   \exp{\left(-\frac{L}{2}\log{\frac{1}{\eta}}\right)}\sqrt{\frac{2\sigma^2\hat{d}}{n}\log\left(\frac{en}{\hat{d}}\right)}~~for~~ n>\hat{d}~.
\end{eqnarray}
\end{theorem}
\begin{theorem} [proved in 8.7]
The binary classification in deep learning is learnable, with the sample complexity of $\widetilde{\mathcal{O}}\left(\sqrt{\frac{\hat{d}}{n}}\right)$\footnote{ We use the notation $\widetilde{\mathcal{O}}$ to hide constants and poly-logarithmic factors of $d$ and $n$.}.
\end{theorem}

\section{Discussions} \label{discuss}
\subsection{Data-fitting and Generalization Trade-off in Deep Learning}

In previous sections, we derived an upper bound on the expected generalization error of deep learning via exploiting the mutual information between the input training set and the output hypothesis. Here, the mutual information quantifies the degree to which the output hypothesis depends on the training data. If the mutual information is small, the output will rely less on the training data, resulting in a better generalization to unseen test data (~i.e. small $G(D, P_{W|S})$~). However, a small mutual information is not helpful to the fitting of training data and thus may lead to a worse expected training error (i.e. large $\mathbb{E}[R_S(W)]$~). 

\begin{figure}
  \centering
    \includegraphics[width=0.4\textwidth]{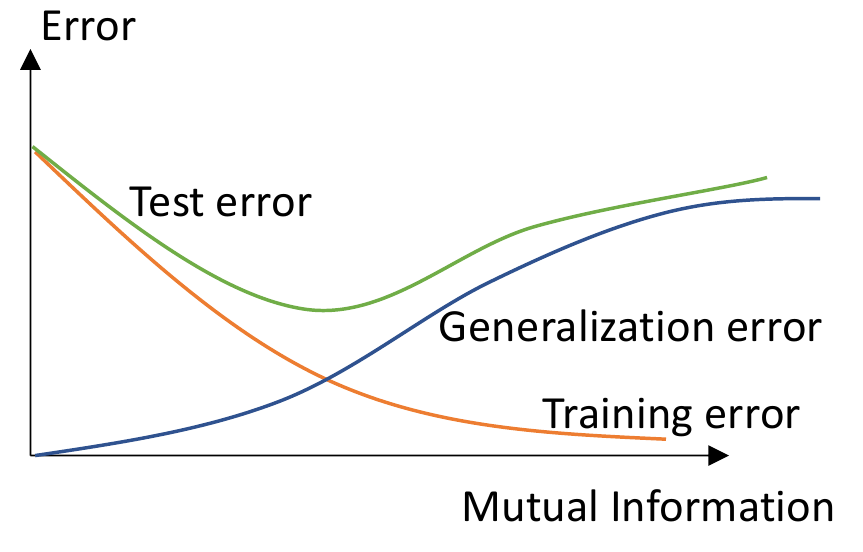}
      \caption{Data-fitting and Generalization Trade-off}
      \label{fig3}
\end{figure}

Our goal is to minimize the expected risk $\mathbb{E}[R(W)]=G(D, P_{W|S}) +\mathbb{E}[R_S(W)]$ in deep learning. In other words, we need to find a right balance between the data fitting and generalization such that the sum of expected generalization error and expected training error is as small as possible. Figure \ref{fig3} illustrates a qualitative relationship between the mutual information and errors of a learning algorithm. Similar to the bias-variance trade-off in traditional learning theory,  we need to control the mutual information between the input and output hypothesis such that the expected risk is small in deep learning.

We have also derived the expected generalization error bound for noisy SGD in deep learning. The injection of Gaussian noise in the updates of the weight $W$ is helpful to reduce the dependence of the output hypothesis on the training data. Therefore, it controls the mutual information and prevents the deep learning algorithm from overfitting. There are also many others ways to control the mutual information between the input and output hypothesis in deep learning, such as dropout and some data augmentation tricks. For example, we can inject noise in the training set (~i.e. $S\to\widetilde{S} $~) and use it to train the deep learning model. By applying the data processing inequality on the Markov chain $S\to\widetilde{S}\to W$, it concludes that the mutual information between the input and output $I(S,W)$ will be smaller and thus achieves a better generalization error. Many other ways for data augmentation can also be interpreted by our theorem, such as random cropping, rotation, and translation of the input images.

\subsection{The Relationship to Algorithmic Hypothesis Class}
The use of mutual information to upper bound the expected generalization error has many advantages in deep learning. It is known that deep models often have unreasonably large parameter space and therefore the predefined hypothesis class of a deep learning model is very large. If we use the complexity of the predefined hypothesis class to upper bound the expected generalization error of deep learning algorithms, the upper bound will be loose. To address this problem, \cite{pmlr-v70-liu17c} introduce the notion of \emph{algorithmic hypothesis class}, which is a subset of the predefined hypothesis class that the learning algorithm is likely to output with high probability. 

As shown in Figure \ref{fig4}, the algorithmic hypothesis class is often much smaller than the predefined hypothesis class because a good learning algorithm always tends to output the hypothesis that fits the input distribution relatively well. Therefore, a tighter generalization bound can be achieved by using the complexity of algorithmic hypothesis class. Our results also adopt the idea of algorithmic hypothesis complexity because the mutual information $I(S,W)$ contains the distribution $P_{W|S}$, which is the subset of predefined hypothesis class that the deep learning algorithm is likely to output.
\begin{figure} \label{fig4}
  \centering
    \includegraphics[width=0.6\textwidth]{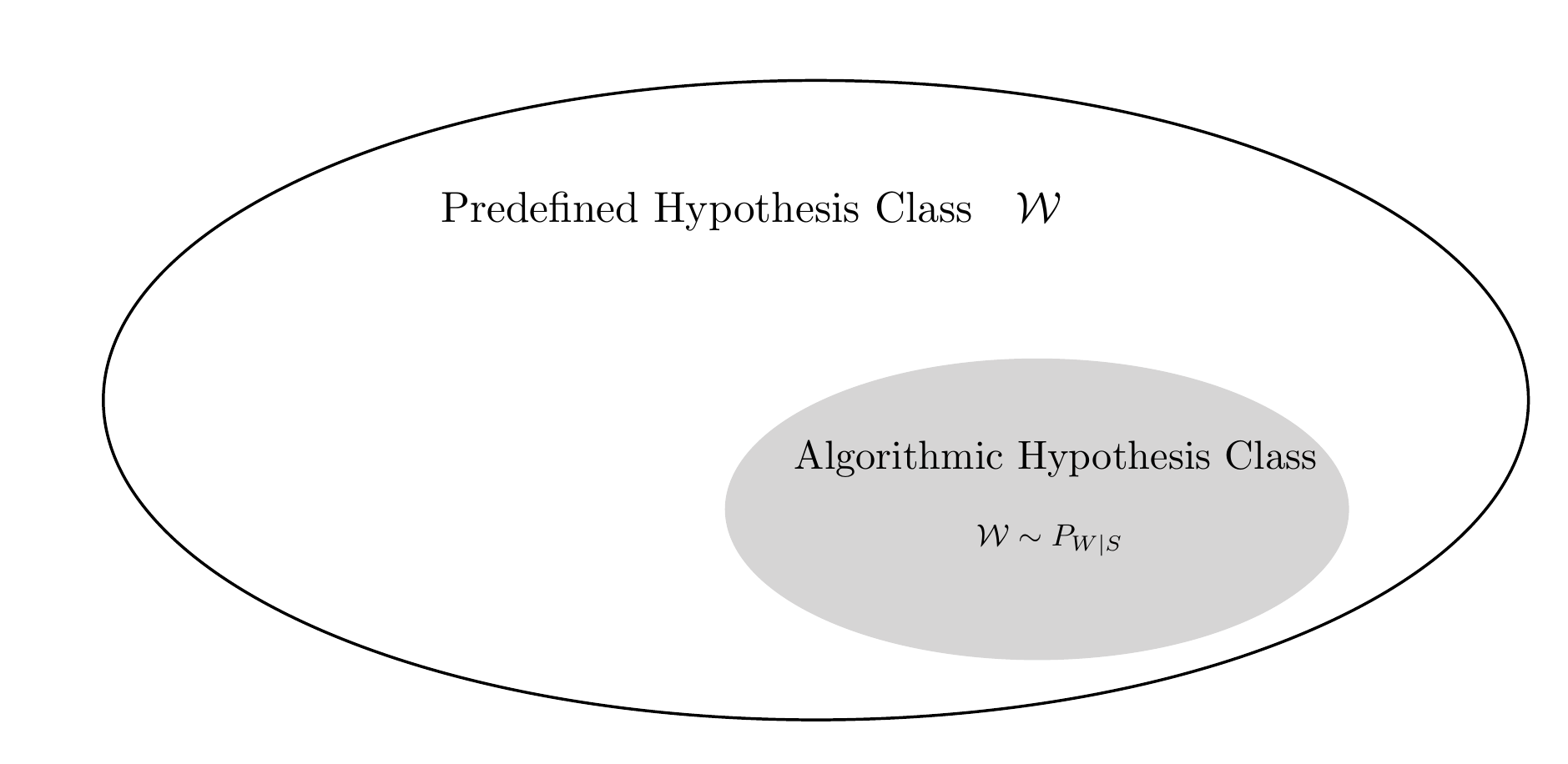}
      \caption{The Implication of Mutual Information on Algorithmic Hypothesis Class}
      \label{fig4}
\end{figure}

\section{Proofs} \label{proofs}

\subsection{Proof of Lemma \ref{lemma2}}

For the $k$-th hidden layer, fixing $w_1,\ldots,w_k$ and considering any input $(x_{k-1},\cdot)\sim D_{k-1}$ and the corresponding output $(x_k,\cdot)\sim P_{D_k|D_{k-1}} $, we have 
\begin{eqnarray}
x_k=\sigma_k(w_kx_{k-1})~.
\end{eqnarray}

Because $rank(w_k)< d_{k-1}$, the dimension of its right null space is greater than or equal to $1$. Denoting the right null space of $w_k$ by $RNULL(w_k)$, then we can pick a non-zero vector $\alpha\in RNULL(w_k)$ such that $w_k\alpha=0$.

Then, we have
\begin{eqnarray}
\sigma_k(w_k(x_{k-1}+\alpha))=\sigma_k(w_kx_{k-1})=x_k~.
\end{eqnarray}
Therefore, for any input $x_{k-1}\sim D_{k-1}$ of the $k$-th hidden layer, there exists $x_{k-1}^{\prime}=x_{k-1}+\alpha$ such that their corresponding outputs are the same. That is, for any $x_{k-1}$, we cannot recover it perfectly.

We conclude that the mapping $P_{D_k|D_{k-1}}$ is noisy and the corresponding layer will cause information loss.

\subsection{Proof of Theorem \ref{main}}

First, by the smoothness of conditional expectation, we have,
 \begin{eqnarray}
&& \mathbb{E}[R(W)-R_S(W)] 
 \nonumber\\
 &&=   \mathbb{E}\left[\mathbb{E}[R(W)-R_S(W)|w_1,\ldots,w_L]\right] ~.
 \end{eqnarray}
We now give an upper bound on $\mathbb{E}[R(W)-R_S(W)|w_1,\ldots,w_L]$.

\begin{lemma} \label{lemma4}
Under the same conditions as in Theorem \ref{main}, the upper bound of $\mathbb{E}[R(W)-R_S(W)|w_1,\ldots,w_L]$ is given by
 \begin{eqnarray}
&&\mathbb{E}[R(W)-R_S(W)|w_1,\ldots,w_L]\nonumber\\
&&\leq \sqrt{\frac{2\sigma^2}{n}I\left(T_L, h|w_1,\ldots,w_L\right)}~.
 \end{eqnarray}
  \begin{proof}
 We have,
 \begin{eqnarray}
&& \mathbb{E}[R(W)-R_S(W)|w_1,\ldots,w_L] \nonumber\\
&&= \mathbb{E}_{h, S}\left[\mathbb{E}_{Z\sim D}[\ell(W, Z)]-\frac{1}{n}\sum_{i=1}^{n}\ell(W,Z_i)|w_1,\ldots,w_L\right]\nonumber
\\&&=\mathbb{E}_{h, T_L}\left[\mathbb{E}_{\widetilde{Z}_L\sim D_L}[\ell(h, \widetilde{Z}_L)]-\frac{1}{n}\sum_{i=1}^{n}\ell(h,Z_{L_i})|w_1,\ldots,w_L\right] ~.
 \end{eqnarray}
We are now going to upper bound
 \begin{eqnarray}
&&\mathbb{E}_{h, T_L}\left[\mathbb{E}_{\widetilde{Z}_L\sim D_L}[\ell(h, \widetilde{Z}_L)]-\frac{1}{n}\sum_{i=1}^{n}\ell(h,Z_{L_i})|w_1,\ldots,w_L \right]~. \nonumber
 \end{eqnarray}
 Note that $T_L\sim D_L^n$ when given $w_1,\dots,w_{L}$, because $T\sim D^n$ and the mappings of hidden layers are given. We adopt the classical idea of ghost sample in statical learning theory. That is, we sample another $T_L^{\prime}$:
\begin{eqnarray}
T_L^{\prime}=\left\{Z_{L_1}^{\prime},\ldots,Z_{L_n}^{\prime} \right\}
\end{eqnarray}
where each element $Z_{L_i}^{\prime}$ is drawn i.i.d. from the distribution $D_L$. We now have,
\begin{eqnarray}
&&\mathbb{E}_{h, T_L}\left[\mathbb{E}_{\widetilde{Z}_L\sim D_L}[\ell(h, \widetilde{Z}_L)]-\frac{1}{n}\sum_{i=1}^{n}\ell(h,Z_{L_i})|w_1,\ldots,w_L\right]\nonumber\\
&&=\mathbb{E}_{h, T_L}\left[ \mathbb{E}_{T_L^{\prime}}\left[\frac{1}{n}\sum_{i=1}^{n}\ell(h,Z_{L_i}^{\prime})\right]-\frac{1}{n}\sum_{i=1}^{n}\ell(h,Z_{L_i})|w_1,\ldots,w_L\right]\nonumber\\
&&=\mathbb{E}_{h, T_L,T_L^{\prime}}\left[\frac{1}{n}\sum_{i=1}^{n}\ell(h,Z_{L_i}^{\prime})|w_1,\ldots,w_L\right]\nonumber\\
&&-\mathbb{E}_{h, T_L}\left[\frac{1}{n}\sum_{i=1}^{n}\ell(h,Z_{L_i})|w_1,\ldots,w_L\right]~.
\end{eqnarray}
We denote the joint distribution of $h$ and $T_L$ by $P_{h, T_L}=P_{h|T_L}\times P_{T_L}$, and the marginal distribution of $h$ and $T_L$ by $P_h$ and $P_{T_L}$ respectively. Therefore, we have,
\begin{eqnarray}
&&\mathbb{E}_{h, T_L,T_L^{\prime}}\left[\frac{1}{n}\sum_{i=1}^{n}\ell(h,Z_{L_i}^{\prime})|w_1,\ldots,w_L\right]-\mathbb{E}_{h, T_L}\left[\frac{1}{n}\sum_{i=1}^{n}\ell(h,Z_{L_i})|w_1,\ldots,w_L\right]\nonumber\\
&&=\mathbb{E}_{h^{\prime}\sim P_h, T_L^{\prime}\sim P_{T_L}}\left[\frac{1}{n}\sum_{i=1}^{n}\ell(h^\prime,Z_{L_i}^{\prime})|w_1,\ldots,w_L\right]-\mathbb{E}_{(h, T_L)\sim P_{h, T_L}}\left[\frac{1}{n}\sum_{i=1}^{n}\ell(h,Z_{L_i})|w_1,\ldots,w_L\right]\nonumber\\
&&=\mathbb{E}_{h^{\prime}\sim P_h, Z_L^{\prime}\sim P_{D_L}}\left[\ell(h^\prime,Z_{L}^{\prime})|w_1,\ldots,w_L\right]-\mathbb{E}_{(h, Z_L)\sim P_{h, D_L}}\left[\ell(h,Z_{L})|w_1,\ldots,w_L\right]~.
\end{eqnarray}

We now bound the above term by the mutual information $I(D_L, h|w_1,\ldots,w_L)$ by employing the following lemma.
\begin{lemma}\cite{donsker1983asymptotic} \label{variational}
Let P and Q be two probability distributions on the same measurable space $\{\Omega,\mathcal{F}\}$. Then the KL-divergence between P and Q can be represented as,
\begin{eqnarray}
D(P||Q)=\sup_{F} \left[\mathbb{E}_P[F]-\log\mathbb{E}_{Q}[e^{F}]\right]
\end{eqnarray}
where the supremum is taken over all measurable functions $F: \Omega\rightarrow \mathbb{R}$ such that $\mathbb{E}_{Q}[e^{F}]<\infty$.
\end{lemma}

Using lemma \ref{variational}, we have,
\begin{eqnarray} \label{flag1}
&& I(D_L, h|w_1,\ldots,w_L)\nonumber\\
&&= D(P_{h, D_L}||P_{h}\times P_{D_L}|w_1,\ldots,w_L)\nonumber\\
&&=\sup_{F} \left[\mathbb{E}_{P_{h, D_L}}\left[F|w_1,\ldots,w_L\right]-\log\mathbb{E}_{P_{h}\times P_{D_L}}\left[e^{F}|w_1,\ldots,w_L\right]\right]\nonumber\\
&&\geq \mathbb{E}_{(h, Z_L)\sim P_{h, D_L}}\left[\lambda\ell(h,Z_{L})|w_1,\ldots,w_L\right]\nonumber\\
&&-\log \mathbb{E}_{h^{\prime}\sim P_h, Z_L^{\prime}\sim P_{D_L}}\left[e^{\lambda\ell(h^\prime,Z_{L_i}^{\prime})}|w_1,\ldots,w_L\right]~.
\end{eqnarray}
As the loss function $\ell(h^{\prime},Z_{L}^{\prime})$ is $\sigma$-sub-Gaussian w.r.t. $(h^{\prime},Z_{L}^{\prime})$, given any $w_1,\ldots,w_L$. By definition, we have,
\begin{eqnarray} \label{flag2}
&&\log \mathbb{E}_{h^{\prime}\sim P_h, Z_L^{\prime}\sim P_{D_L}}\left[e^{\lambda\ell(h^\prime,Z_{L}^{\prime})}|w_1,\ldots,w_L\right] \nonumber\\
&&  \leq \frac{\sigma^2\lambda^2}{2}+\mathbb{E}_{h^{\prime}\sim P_h, Z_L^{\prime}\sim P_{D_L}}\left[\lambda\ell(h^\prime,Z_{L}^{\prime})|w_1,\ldots,w_L\right]~.
\end{eqnarray}
Substituting inequality (\ref{flag2}) into inequality (\ref{flag1}), we have,
\begin{eqnarray} \label{flag3}
&&  \mathbb{E}_{(h, Z_L)\sim P_{h, D_L}}\left[\lambda\ell(h,Z_{L})|w_1,\ldots,w_L\right]-\frac{\sigma^2\lambda^2}{2}-\nonumber\\
&&\mathbb{E}_{h^{\prime}\sim P_h, Z_L^{\prime}\sim P_{D_L}}\left[\lambda\ell(h^\prime,Z_{L}^{\prime})|w_1,\ldots,w_L\right]- I(D_L, h|w_1,\ldots,w_L)\nonumber\\
&&=-\frac{\sigma^2\lambda^2}{2}+\left[\mathbb{E}_{(h, Z_L)\sim P_{h, D_L}}\left[\ell(h,Z_{L})|w_1,\ldots,w_L\right]\right.\nonumber\\
&&\left.-\mathbb{E}_{h^{\prime}\sim P_h, Z_L^{\prime}\sim P_{D_L}}\left[\ell(h^\prime,Z_{L}^{\prime})|w_1,\ldots,w_L\right]\right]\lambda - I(D_L, h|w_1,\ldots,w_L)\leq 0.
\end{eqnarray}
The left side of the above inequality is a quadratic curve about $\lambda$ and is always less than or equal to zero. Therefore we have,
\begin{eqnarray} \label{flag4}
&&\left|\mathbb{E}_{(h, Z_L)\sim P_{h, D_L}}\left[\ell(h,Z_{L})|w_1,\ldots,w_L\right]-\mathbb{E}_{h^{\prime}\sim P_h, Z_L^{\prime}\sim P_{D_L}}\left[\ell(h^\prime,Z_{L}^{\prime})|w_1,\ldots,w_L\right]\right|^2 \nonumber\\
&&\leq 2\sigma^2I(D_L, h|w_1,\ldots,w_L)~.
\end{eqnarray}
As $T_L\sim D_L^n$ given $w_1,\ldots,w_L$, we have,\footnote{Here, the conditional entropy $H_{cond}(Z_{L_1},..,\ldots,Z_{L_n}|w_1,\ldots,w_L;h)$ and $H_{cond}(Z_{L_i}|w_1,\ldots,w_L;h,Z_{L_{i-1}},\ldots,Z_{L_1})$ only take expectation over $h$ and $(h,Z_{L_{i-1}},\ldots,Z_{L_1})$ respectively. In other words, $w_1,\ldots,w_L$ are given.}
\begin{eqnarray}
&&I(T_L,h|w_1,\ldots,w_L)\nonumber\\
&& = I(Z_{L_1},..,\ldots,Z_{L_n};h|w_1,\ldots,w_L)\nonumber\\
&& = H(Z_{L_1},..,\ldots,Z_{L_n}|w_1,\ldots,w_L)-H_{cond}(Z_{L_1},..,\ldots,Z_{L_n}|w_1,\ldots,w_L;h)  \nonumber\\
&&=\sum_{i=1}^{n} H(Z_{L_i}|w_1,\ldots,w_L)-\sum_{i=1}^{n} H_{cond}(Z_{L_i}|w_1,\ldots,w_L;h,Z_{L_{i-1}},\ldots,Z_{L_1})\nonumber\\
&&\geq \sum_{i=1}^{n} H(Z_{L_i}|w_1,\ldots,w_L)-\sum_{i=1}^{n} H_{cond}(Z_{L_i}|w_1,\ldots,w_L;h)\nonumber\\
&&=nI(Z_L;h|w_1,\ldots,w_L)\nonumber\\
&&=nI(D_L;h|w_1,\ldots,w_L)~.
\end{eqnarray}
In other words, we have
\begin{eqnarray} \label{eq44}
I(D_L;h|w_1,\ldots,w_L)\leq \frac{I(T_L,h|w_1,\ldots,w_L)}{n}~.
\end{eqnarray}
We finish the proof by substituting (\ref{eq44}) into (\ref{flag4}).
\end{proof}
   
\end{lemma}
  By Theorem \ref{thm1}, we can use the strong data processing inequality for the Markov chain in Figure \ref{fig2} recursively. Thus, we have,
 \begin{eqnarray} \label{flag5}
&&\sqrt{\frac{2\sigma^2}{n}I\left(T_L, h|w_1,\ldots,w_L\right)}\nonumber\\
&&\leq \sqrt{\frac{2\sigma^2}{n}\eta_{L}I\left(T_{L-1}, h|w_1,\ldots,w_L\right)}\nonumber
\\
&&\leq \sqrt{\frac{2\sigma^2}{n}\eta_{L}\eta_{L-1}I\left(T_{L-2}, h|w_1,\ldots,w_L\right)}\nonumber\\
&&
\leq\ldots\leq \sqrt{\frac{2\sigma^2}{n}
\left(\prod_{k=1}^{L}\eta_{k}\right)I\left(S, h|w_1,\ldots,w_L\right)} ~.
 \end{eqnarray}
 
We then have
 \begin{eqnarray} \label{note}
&& \mathbb{E}[R(W)-R_S(W)] \nonumber \\ && 
 =   \mathbb{E}\left[\mathbb{E}[R(W)-R_S(W)|w_1,\ldots,w_L]\right]  \nonumber\\
 &&
 \leq \mathbb{E}_{w_1,\ldots,w_L}\left(\sqrt{\frac{2\sigma^2}{n}
\left(\prod_{k=1}^{L}\eta_{k}\right)I\left(S, h|w_1,\ldots,w_L\right)}\right)\nonumber\\
 &&
= \mathbb{E}_{w_1,\ldots,w_L}\left(\sqrt{\prod_{k=1}^{L}\eta_{k}}\sqrt{\frac{2\sigma^2}{n}I(S,(w_1,\ldots,w_L,h)|w_1,\ldots,w_L) }\right) \nonumber\\
 &&
= \mathbb{E}_{w_1,\ldots,w_L}\left(\sqrt{\prod_{k=1}^{L}\eta_{k}}\sqrt{\frac{2\sigma^2}{n}I(S,W|w_1,\ldots,w_L) }\right)  \nonumber\\
 &&\leq\sqrt{\mathbb{E}_{w_1,\ldots,w_L}\left(\prod_{k=1}^{L}\eta_{k}\right)}\sqrt{\frac{2\sigma^2}{n}\mathbb{E}_{w_1,\ldots,w_L}[I(S,W|w_1,\ldots,w_L)] }
 \end{eqnarray}
 
 As conditions reduce the entropy, we have the following relations
 \begin{eqnarray}
&& \mathbb{E}_{w_1,\ldots,w_L} \left[I(S,W | w_1,\ldots,w_L)\right]\nonumber\\ 
&& =I_{cond}(S,W | w_1,\ldots,w_L) \nonumber\\ 
&& =H_{cond}(S|w_1,\ldots,w_L) - H_{cond}(S|W,w_1,\ldots,w_L)\nonumber\\ 
&&\leq H(S) - H_{cond}(S|W) = I(S,W)~.
\end{eqnarray}

Therefore, we have
 \begin{eqnarray} \label{eq48}
 &&\sqrt{\mathbb{E}_{w_1,\ldots,w_L}\left(\prod_{k=1}^{L}\eta_{k}\right)}\sqrt{\frac{2\sigma^2}{n}\mathbb{E}_{w_1,\ldots,w_L}[I(S,W|w_1,\ldots,w_L)] }
 \nonumber\\
 &&\leq \sqrt{\mathbb{E}_{w_1,\ldots,w_L}\left(\prod_{k=1}^{L}\eta_{k}\right)}\sqrt{\frac{2\sigma^2}{n}I(S,W) }\nonumber\\
 &&= \sqrt{\eta^L}\sqrt{\frac{2\sigma^2}{n}I(S,W) }\nonumber\\
 &&=  \exp{\left(-\frac{L}{2}\log{\frac{1}{\eta}}\right)}\sqrt{\frac{2\sigma^2}{n}I(S,W) }~.
  \end{eqnarray}

where  \begin{eqnarray}
 \eta=\left(\mathbb{E}_{w_1,\ldots,w_L}\left(\prod_{k=1}^{L}\eta_k\right)\right)^{\frac{1}{L}} < 1 ~.
 \end{eqnarray}
 It's worth mentioning that the information loss factor $\eta_k$ has been taken an expectation w.r.t. the weight $w_1,\ldots,w_L$, which further implies that the information loss can be applied to the fully connected layer $w_k$ even with $d_k\geq d_{k-1}$, as long as $rank(w_k)<d_{k-1}$ holds for some value of $w_1,\ldots,w_L$ with non-zero probability.

\subsection{Proof of Theorem \ref{thm3}}
Let $S'=(Z_1^{\prime},\cdots,Z_n^{\prime})$ be a ghost sample of $S$. We have
\begin{eqnarray}\label{eq1}
&&\mathbb{E}[R(W)-R_S(W)]\nonumber\\
&&=\mathbb{E}_{W\sim P_{W|S}}\left[\mathbb{E}_{S,S'}\left[\frac{1}{n}\sum_{i=1}^{n}\ell(W,Z_i^{\prime})\right]-\mathbb{E}_{S}\left[\frac{1}{n}\sum_{i=1}^{n}\ell(W,{Z}_i)\right]\right]\nonumber\\
&&=\mathbb{E}_{W\sim P_{W|S}}\left[\frac{1}{n}\sum_{i=1}^{n}\mathbb{E}_{S,Z_i^{\prime}}\left[\ell(W,Z_i^{\prime})\right]-\frac{1}{n}\sum_{i=1}^{n}\mathbb{E}_{S}\left[\ell(W,{Z}_i)\right]\right]\nonumber\\ 
&&=\mathbb{E}_{W\sim P_{W|S}}\left[\frac{1}{n}\sum_{i=1}^{n}\mathbb{E}_{S,Z_i^{\prime}}\left[\ell(W,Z_i^{\prime})\right]-\frac{1}{n}\sum_{i=1}^{n}\mathbb{E}_{S,Z_i^{\prime},W^i\sim P_{W^i|S^i}}\left[\ell(W^i,Z_i^{\prime})\right]\right]\nonumber\\
&&=\frac{1}{n}\sum_{i=1}^{n}\mathbb{E}_{S,Z_i^{\prime},W\sim P_{W|S}, W^i\sim P_{W^i|S^i}}\left[\ell(W,Z_i^{\prime})-\ell(W^i,Z_i^{\prime})\right],
\end{eqnarray}
where $W^i$ stands for the output of the learning algorithm when the input is $S^i=(Z_1, \cdots, Z_{i-1}, Z_i^{\prime},$ $Z_{i+1}, \cdots, Z_n)$ and $Z_i^{\prime}$, and $Z_i$ ($i=1,\ldots,n$) are i.i.d. examples.\\
From equation (\ref{flag4}) and (\ref{eq44}), we have
\begin{eqnarray}
&&\left|\mathbb{E}[R(W)-R_S(W)|w_1,\ldots,w_L]\right| \leq \sqrt{ \frac{2\sigma^2}{n}I\left(T_L, h|w_1,\ldots,w_L\right)}~.
\end{eqnarray}
Using similar proofs as in Theorem \ref{main}, we have  
 \begin{eqnarray} \label{eq2}
&&\left| \mathbb{E}[R(W)-R_S(W)] \right | \leq \exp{\left(-\frac{L}{2}\log{\frac{1}{\eta}}\right)}\sqrt{\frac{2\sigma^2}{n}I(S,W) }~.
 \end{eqnarray}
Note that the difference between the above inequality and our main theorem is that the absolute value is adopted for the expected generalization error, which may be slightly tighter, but the conclusions are almost the same.
Combining (\ref{eq1}) and (\ref{eq2}), we have
\begin{eqnarray}
&&\left | \frac{1}{n}\sum_{i=1}^{n}\mathbb{E}_{S\sim D^n,Z_i^{\prime}\sim D, W\sim P_{W|S}, W^i\sim P_{W^i|S^i}}\left[\ell(W,Z_i^{\prime})-\ell(W^i,Z_i^{\prime})\right]\right| \nonumber \\&& \leq \exp{\left(-\frac{L}{2}\log{\frac{1}{\eta}}\right)}\sqrt{\frac{2\sigma^2}{n}I(S,W) }
\end{eqnarray}
which ends the proof.

\subsection{Proof of Theorem \ref{SGD}}
By (\ref{flag4}), (\ref{eq44}), and ({\ref{flag5}}), we have,
\begin{eqnarray}   \label{flag6}
&& \left|\mathbb{E}[R(W)-R_S(W)|w_1,\ldots,w_L]\right|  \leq \sqrt{\prod_{k=1}^{L}\eta_{k}}\sqrt{\frac{2\sigma^2}{n}I(S,h|w_1,\ldots,w_L) } ~.
\end{eqnarray}
We now bound the expectation of the right side of the above inequality and use the smoothness of conditional expectation. Then the theorem can be proved. Our analysis here is based on the work of \cite{raginsky2016information} and \cite{2018arXiv180104295P}. 
 
At the final iteration, we have $t=T$ and the algorithm outputs $W=W_T$. We have the following Markov relationship when given $w_1,\ldots,w_L$ and the initialization $W_0$ is known,
\begin{eqnarray}
&&D\rightarrow S \rightarrow [\bold{Z}_1,\ldots,\bold{Z}_T] \rightarrow [W_1, \ldots, W_T]\rightarrow [h_1,\ldots,h_T] \rightarrow h_T~.
\end{eqnarray}
Therefore, we have,
\begin{eqnarray} \label{flag13}
&&\mathbb{E}_{w_1,\ldots,w_L}[I(S, h|w_1,\ldots,w_L)]\nonumber\\
&&=I_{cond}(S, h|w_1,\ldots,w_L)\nonumber\\
&&= I_{cond}(S, h_T|w_1,\ldots,w_L)\nonumber\\
&&\leq I_{cond}(S; h_1, \ldots, h_T|w_1,\ldots,w_L) \nonumber\\
&& \leq I_{cond}(\bold{Z}_1,\ldots, \bold{Z}_T;  h_1, \ldots, h_T|w_1,\ldots,w_L)~.
\end{eqnarray}
Using the chain rule of mutual information, we have
\begin{eqnarray} \label{flag12}
&&I_{cond}(\bold{Z}_1,\ldots, \bold{Z}_T;  h_1, \ldots, h_T|w_1,\ldots,w_L)=\nonumber\\
&&\sum_{i=1}^{T}I_{cond}(\bold{Z}_1,\ldots, \bold{Z}_T; h_i | h_{i-1},\ldots, h_1; w_1,\ldots,w_L ).
\end{eqnarray}

By definition, we have
\begin{eqnarray} \label{flag10}
&& I_{cond}(\bold{Z}_1,\ldots, \bold{Z}_T; h_i | h_{i-1},\ldots, h_1; w_1,\ldots,w_L )\nonumber \\
&&= H_{cond}(h_i | h_{i-1},\ldots, h_1; w_1,\ldots,w_L)- H_{cond}(h_i |\bold{Z}_1,\ldots, \bold{Z}_T; h_{i-1},\ldots, h_1; w_1,\ldots,w_L)\nonumber \\
&&=H_{cond}(h_i | h_{i-1}; w_1,\ldots,w_L) - H_{cond}(h_i | h_{i-1}; \bold{Z}_{i};w_1,\ldots,w_L) \nonumber \\
&&\leq H_{cond}(h_i | h_{i-1})  - H_{cond}\left(h_{i-1} - \alpha_i\left[\frac{1}{m}\sum_{j=1}^{m}\nabla_{h}\ell(W_{i-1}, Z_{i_j})\right]+n_i | h_{i-1}; \bold{Z}_{i}; w_1,\ldots,w_L\right) \nonumber \\
&&= H_{cond}(h_i | h_{i-1})    - H_{cond}(n_i | h_{i-1}; \bold{Z}_{i} ) \nonumber \\
&&= H_{cond}(h_i - h_{i-1} | h_{i-1})  - H(n_i )
\end{eqnarray}
where the last equality follows from the fact that translation does not affect the entropy of a random variable. From updating rules, we have
\begin{eqnarray}
&& \mathbb{E}\left[||h_i - h_{i-1}||^2 \right]  \nonumber\\
&&= \mathbb{E}\left[ \left|\left|  - \alpha_i\left[\frac{1}{m}\sum_{j=1}^{m}\nabla_{h}\ell(W_{i-1}, Z_{i_j})\right]+n_i \right|\right|^2   \right]  \nonumber\\
&&= \mathbb{E}\left[ \left|\left|  - \alpha_i\left[\frac{1}{m}\sum_{j=1}^{m}\nabla_{h}\ell(W_{i-1}, Z_{i_j})\right] \right|\right|^2 +   \left|\left| n_i \right|\right|^2 \right]           \nonumber \\
&& \leq M^2\alpha_i^2 +d\sigma_i^2~.
\end{eqnarray}
It is known that for a random variable with constraints up to the second moment, the Gaussian distribution reaches the maximum entropy \cite{cover2012elements}. For a Gaussian random variable $X \sim \mathcal{N}(0, \sigma^2 \bold{I}_d)$ , the entropy of $X$ is 
\begin{equation}
H(X) = \frac{d}{2}\log(2\pi e\sigma^2)~.
\end{equation}
Therefore, we have,
\begin{eqnarray} \label{flag8}
&&H_{cond}(h_i - h_{i-1} | h_{i-1})\leq  \frac{d}{2}\log\left(2\pi e\left(\frac{\alpha_i^2M^2}{d}+\sigma_i^2\right)\right)~.
\end{eqnarray}
We also have,
\begin{equation} \label{flag9}
 H(n_i )= \frac{d}{2}\log(2\pi e \sigma_i^2)~.
\end{equation}
Substituting (\ref{flag8}) and (\ref{flag9}) into (\ref{flag10}), we have,
\begin{eqnarray} \label{flag11}
&& I_{cond}(\bold{Z}_1,\ldots, \bold{Z}_T; h_i | h_{i-1},\ldots, h_1; w_1,\ldots,w_L ) \nonumber\\
&&\leq  \frac{d}{2}\log\left(2\pi e\left(\frac{\alpha_i^2M^2}{d}+\sigma_i^2\right)\right)-\frac{d}{2}\log(2\pi e \sigma_i^2)\nonumber \\
&& = \frac{d}{2}\log\left(\frac{\alpha_i^2M^2}{d\sigma_i^2}+1\right) \leq \frac{\alpha_i^2M^2}{2\sigma_i^2}~.
\end{eqnarray}

By (\ref{flag6}), (\ref{flag13}), (\ref{flag12}), and (\ref{flag11}), similar to the steps of (\ref{note}) and (\ref{eq48}), we have
\begin{eqnarray}
&& \left|\mathbb{E}[R(W)-R_S(W)]\right| \nonumber \\
&&= \left|\mathbb{E}\left[\mathbb{E}\left[R(W)-R_S(W)|w_1,\ldots,w_L\right]\right]\right| \nonumber \\
&&\leq \left| \sqrt{\mathbb{E}\left[ \prod_{k=1}^{L}\eta_{k} \right]}\sqrt{\frac{2\sigma^2}{n}I_{cond}(S,h|w_1,\ldots,w_L) } \right|\nonumber \\
&& \leq\exp{\left(-\frac{L}{2}\log{\frac{1}{\eta}}\right)}\sqrt{\frac{\sigma^2}{n}\sum_{i=1}^{T}\frac{M^2\alpha_i^2}{\sigma_i^2} }
\end{eqnarray}
which completes the proof.

\subsection{Proof of Theorem \ref{BINARY}}
 
From (\ref{flag6}), we have
\begin{eqnarray}   \label{flag14}
&& \left|\mathbb{E}[R(W)-R_S(W)|w_1,\ldots,w_L]\right| \nonumber \\
&& \leq  \sqrt{\prod_{k=1}^{L}\eta_{k}}\sqrt{\frac{2\sigma^2}{n}I(S,h|w_1,\ldots,w_L) }\nonumber  \\
&& \leq \sqrt{\prod_{k=1}^{L}\eta_{k}}\sqrt{\frac{2\sigma^2}{n}H(\mathcal{H}) }\nonumber  \\
&& \leq  \sqrt{\prod_{k=1}^{L}\eta_{k}}\sqrt{\frac{2\sigma^2}{n}\log\Pi(n) }~.
\end{eqnarray}
Using Sauer's Lemma presented in \cite{mohri2012foundations}, we have
\begin{eqnarray}  \label{sauer}
&&\Pi(n)\leq 2^n\leq 2^{\hat{d}} \quad for \quad n\leq \hat{d} \label{sauer1}~, \\
&& \Pi(n)\leq \left(\frac{en}{\hat{d}}\right)^{\hat{d}}\quad for \quad n> \hat{d}  \label{sauer2}~.
\end{eqnarray}
Substituting (\ref{sauer1}) and (\ref{sauer2}) into (\ref{flag14}), and using the smoothness of conditional expectation as in (\ref{note}), one can complete the proof.

 \subsection{Learnability of Noisy SGD for Deep Learning}
For the case of noisy SGD in deep learning, following Theorem \ref{SGD} and similar steps of deriving equation (\ref{eqc}), we have
\begin{equation} 
\mathbb{E}_{W, S}[R(W)] - R^* \leq  \exp{\left(-\frac{L}{2}\log{\frac{1}{\eta}}\right)}\sqrt{\frac{\sigma^2}{n}\sum_{i=1}^{T}\frac{M^2\alpha_i^2}{\sigma_i^2} }~. 
\end{equation}
Similar to previous proofs, by Markov inequality, we have that with probability at least $1-\delta$
\begin{equation} 
R(W) - R^* \leq \frac{1}{\delta}\exp{\left(-\frac{L}{2}\log{\frac{1}{\eta}}\right)}\sqrt{\frac{\sigma^2}{n}\sum_{i=1}^{T}\frac{M^2\alpha_i^2}{\sigma_i^2} }~. 
\end{equation}
If we set the learning rate and the variance of noise as $\alpha_i= C/i^2 $ and $\sigma_i=\sqrt{C/i^2}$, we have that, with probability at least $1-\delta$,
\begin{eqnarray} 
&&R(W) - R^* \leq \frac{1}{\delta}\exp{\left(-\frac{L}{2}\log{\frac{1}{\eta}}\right)}\sqrt{\frac{CM^2\sigma^2}{n}\sum_{i=1}^{T}\frac{1}{i^2} }\nonumber \\
&&\leq  \frac{1}{\delta}\exp{\left(-\frac{L}{2}\log{\frac{1}{\eta}}\right)}\sqrt{\frac{CM^2\sigma^2\pi^2}{6n} }
\end{eqnarray}
where the last inequality is obtained by the analytic continuation of Riemann zeta function $\zeta(s)$ at $s=2$,
\begin{eqnarray}
\sum_{i=1}^{T}\frac{1}{i^2}\leq \sum_{i=1}^{\infty}\frac{1}{i^2} =\zeta(2) =\frac{\pi^2}{6}~.
\end{eqnarray}

In this case, the noisy SGD in deep learning is learnable with sample complexity $\mathcal{O}\left(\frac{1}{\sqrt{n}}\right)$.

\subsection{ Learnability of Binary Classification in Deep Learning}
Similar to the case of noisy SGD, based on Theorem \ref{BINARY} and following similar steps of deriving equation (\ref{eqc}) and using Markov inequality, we have that with probability at least $1-\delta$,
\begin{eqnarray} 
R(W) - R^* \leq \frac{1}{\delta} \exp{\left(-\frac{L}{2}\log{\frac{1}{\eta}}\right)}\sqrt{\frac{2\sigma^2\hat{d}}{n}}\quad for\quad n\leq \hat{d}
\end{eqnarray}
and
\begin{eqnarray} 
&& R(W) - R^* \leq \frac{1}{\delta} \exp{\left(-\frac{L}{2}\log{\frac{1}{\eta}}\right)}\sqrt{\frac{2\sigma^2\hat{d}}{n}\log\left(\frac{en}{\hat{d}}\right)}~~~for\quad n>\hat{d}~.
\end{eqnarray}
Therefore, the binary classification in deep learning is learnable with sample complexity $\widetilde{\mathcal{O}}\left(\sqrt{\frac{\hat{d}}{n}}\right)$~.

\section{Conclusions}\label{section5}
In this paper, we obtain an exponential-type upper bound for the expected generalization error of deep learning and prove that deep learning satisfies a weak notion of stability. Besides, we also prove that deep learning algorithms are learnable in some specific cases such as employing noisy SGD and for binary classification. Our results have valuable implications for other critical problems in deep learning that require further investigation. (1) Traditional statistical learning theory can validate the success of deep neural networks, because (i) the mutual information between the learned feature and classifier $I(T_L, h|w_1,\ldots,w_L)$ decreases with increasing number of contraction layers $L$, and (ii) smaller mutual information implies higher algorithmic stability \cite{raginsky2016information} and smaller complexity of the algorithmic hypothesis class \cite{pmlr-v70-liu17c}. (2) The information loss factor $\eta< 1$ offers the potential to explore the characteristics of various convolution, pooling, and activation functions as well as other deep learning tricks; that is, how they contribute to the reduction in the expected generalization error. (3) The weak notion of stability for deep learning is only a necessary condition for learnability \cite{shalev2010learnability} and deep learning is learnable in some specific cases. It would be interesting to explore a necessary \emph{and} sufficient condition for the learnability of deep learning in a general setting. (4) When increasing the number of contraction layers in DNNs, it is worth further exploring: how to filter out redundant information while keep the useful part intact.

\bibliographystyle{apalike}
\bibliography{arxiv}

\end{document}